\documentclass[10pt,twocolumn,letterpaper]{article}

\usepackage{cvpr}
\usepackage{times}
\usepackage{epsfig}
\usepackage{times}  
\usepackage{helvet}  
\usepackage{courier}  
\usepackage{url}  
\usepackage{graphicx}  
\usepackage{subfigure}
\usepackage{amsmath}
\usepackage{amssymb}
\usepackage{bbm}
\usepackage{wasysym}
\usepackage{xcolor}
\usepackage{colortbl,booktabs}
\usepackage{threeparttable}
\usepackage{multirow}
\usepackage{multicol}
\newcommand{\rulev}{\unskip\ \vrule\ }
\newcommand{\ruleh}{\unskip\ \hrule\ }

\usepackage[breaklinks=true,bookmarks=false]{hyperref}

\cvprfinalcopy 

\def\cvprPaperID{****} 

\begin{document}

\title{Chinese Typeface Transformation with Hierarchical Adversarial Network}

\author{Jie Chang \quad Yujun Gu\quad  Ya Zhang\\
Cooperative Madianet Innovation Center\\
Shanghai Jiao Tong University\\
{\tt\small \{j\_chang, yjgu, ya\_zhang\}@sjtu.edu.cn}
}
\maketitle

\begin{abstract}
In this paper, we explore automated typeface generation through image style transfer which has shown great promise in natural image generation. Existing style transfer methods for natural images generally assume that the source and target images share similar high-frequency features. However, this assumption is no longer true in typeface transformation. Inspired by the recent advancement in Generative Adversarial Networks (GANs), we propose a Hierarchical Adversarial Network (HAN) for typeface transformation. The proposed HAN consists of two sub-networks: a transfer network and a hierarchical adversarial discriminator. The transfer network maps characters from one typeface to another. A unique characteristic of typefaces is that the same radicals may have quite different appearances in different characters even under the same typeface. Hence, a stage-decoder is employed by the transfer network to leverage multiple feature layers, aiming to capture both the global and local features. The hierarchical adversarial discriminator implicitly measures data discrepancy between the generated domain and the target domain. To leverage the complementary discriminating capability of different feature layers, a hierarchical structure is proposed for the discriminator. We have experimentally demonstrated that HAN is an effective framework for typeface transfer and characters restoration.
\end{abstract}

\section{Introduction}
\label{introduction}
Chinese Typeface design is a very time-consuming task, requiring considerable efforts on manual design of benchmark characters. Automated typeface synthesis, i.e. synthesizing characters of a certain typeface given few manually designed samples, has been explored, usually based on manually extracted features. For example, each Chinese character is treated as a combination of its radicals and strokes, and shape representation of specified typefaces such as the contour, orientation and the component size are explicitly learned ~\cite{Xu2009Automatic,Xu2010Automatic,Zhou2011Easy,Zhang2016Drawing,Xiao2016Automatic}. However, these manual features heavily relies on preceding structural segmentation of characters, which itself is a non-trivial task and heavily affected by prior knowledge.

\begin{figure}[tbp]
\centering
\includegraphics[width=0.47\textwidth]{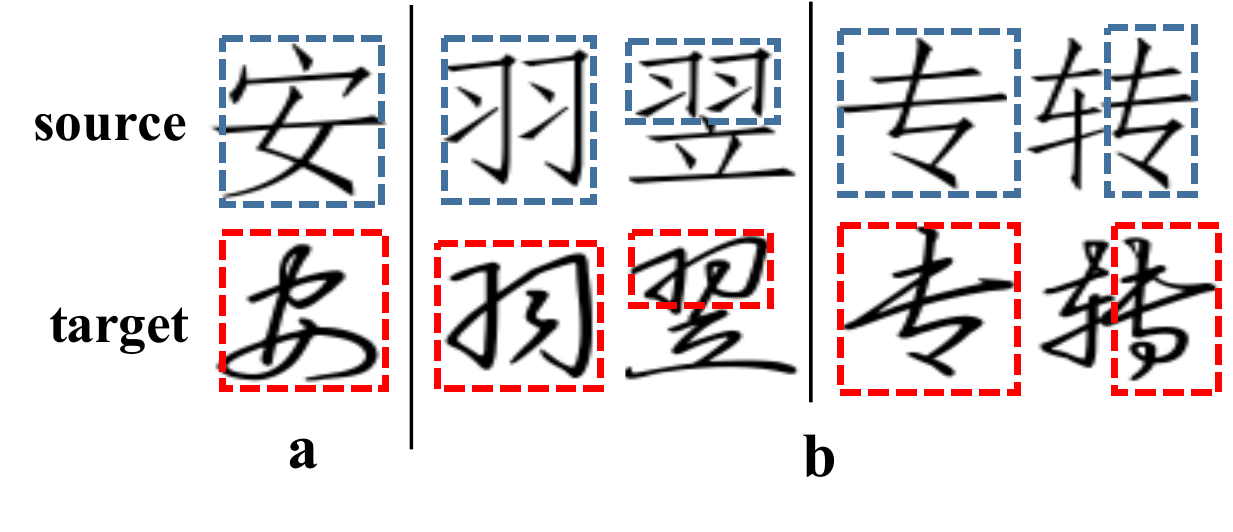}
\caption{(a) target style twists strokes in source character, making they do not share the invariant high-frequency features though they are the same character semantically. (b)
The components in blue dotted box share the same radicals but their corresponding ones (in red dotted box) with target style are quite different.}
\label{fig: property}
\end{figure}

In this paper, we model typeface transformation as an image-to-image transformation problem and attempt to directly learn the transformation  end-to-end. Typically, image-to-image transformation involves a transfer network to map the source images to target images. A set of losses are proposed in learning the transfer network. The pixel loss is defined as pixel-wise difference between the output and the corresponding ground-truth~\cite{Long2014Fully,Isola2016Image}. The perceptual loss~\cite{Johnson2016Perceptual}, perceptual similarity~\cite{Dosovitskiy2016Generating} and style\&content loss~\cite{Chen2017StyleBank} are proposed to evaluate the differences between hidden-level features and all are based on the ideology of feature matching~\cite{Salimans2016Improved}.
More recently, several variant of generative adversarial networks (e.g CGAN~\cite{Mirza2014Conditional}, CycleGAN~\cite{Zhu2017Unpaired}), which introduce a discriminant network in addition to the transfer network for adversarial learning, have been successfully applied to image-to-image transformation including in-painting~\cite{Pathak2016Context}, de-noising~\cite{Zhang2017Image} and super-resolution~\cite{Ledig2016Photo}. While the above methods have shown great promise for various applications, they are not directly applicable to typeface transformation due to the following domain specific characteristics.
\begin{itemize}
\setlength{\itemsep}{0pt}
\setlength{\parsep}{0pt}
\setlength{\parskip}{0pt}
\item  Different from style-transfer between natural images where the source image shares high-frequency features with the target image, the transformation between two different typefaces usually leads to distortion of strokes or radicals (e.g Fig~\ref{fig: property}),  meaning change between different styles leads to change of high-level representations. Hence, we cannot use a pre-trained network(e.g. VGG~\cite{Simonyan2014Very}) to extract high-level representations as invariant content representation in training or explicitly define the style representation.
\item For typeface transformation task, different characters may share the same radicals. This is a nice peculiarity that typeface transformation methods can leverage, i.e. learning the direct mapping of radicals between source and target styles. However, sometimes in one certain typeface, the same radicals may appear quite differently in different characters. Fig~\ref{fig: property}(b) presents two examples where certain radicals have different appearance in another styles. It will leads to severe over-fitting if we just considering the global property while ignore detailed local information.
\end{itemize}

\begin{figure*}[hptb]
\centering
\includegraphics[width=0.99\textwidth]{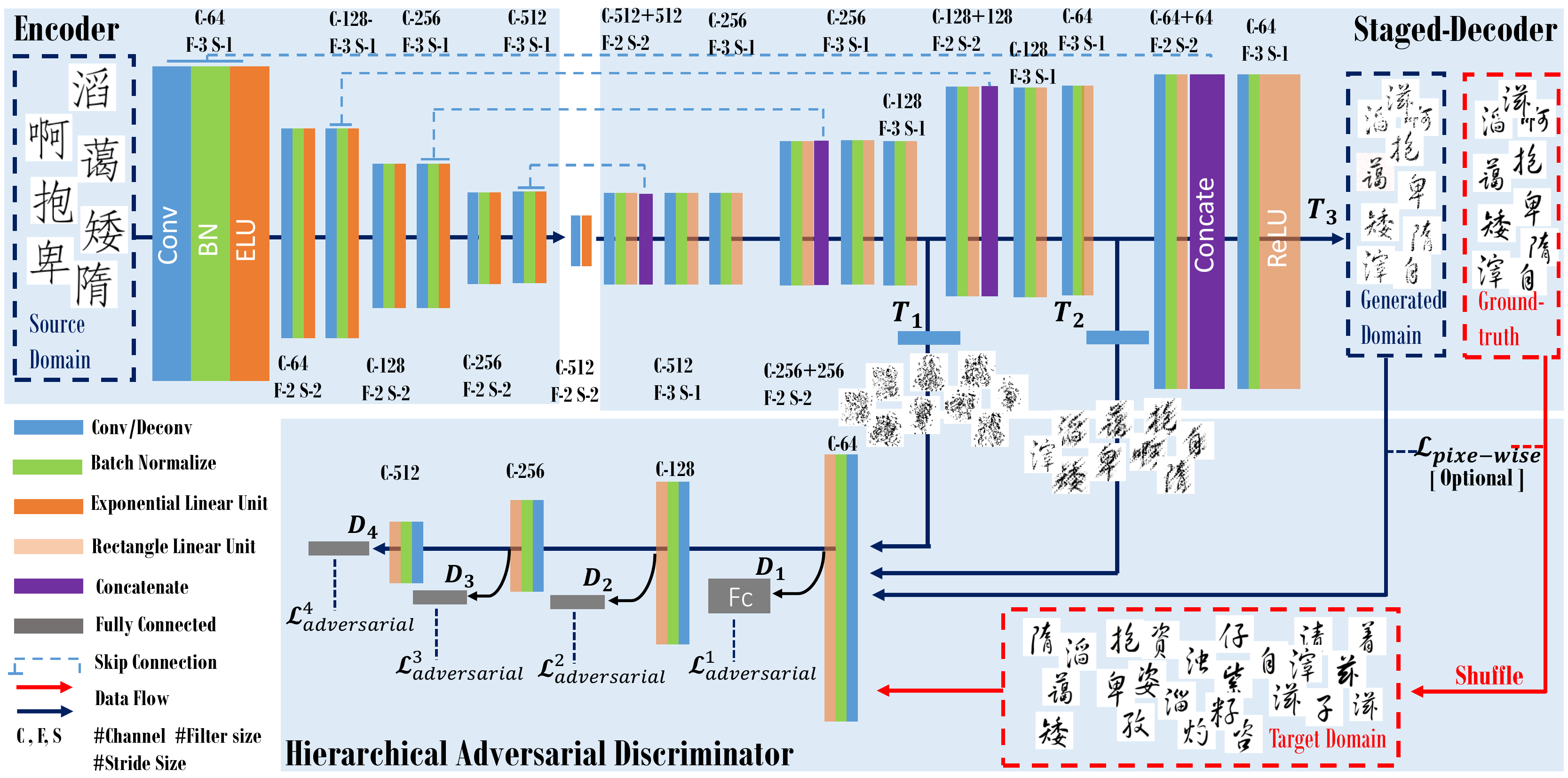}
\caption{The proposed Hierarchical Adversarial Network(HAN). HAN consists of an Encoder, a Staged-Decoder and a Hierarchical Adversarial Discriminator. The Encoder follows the Conv-BatchNorm~\cite{Ioffe2015Batch}-ELU~\cite{Clevert2015Fast} architecture. The Staged-Decoder follows the Conv-BatchNorm-ReLU while two extra transformed characters are decoded from two intermediate features. The hierarchical adversarial discriminator is used to distinguish the transformed characters and the ground-truth from multi-level features.}
\label{fig: HierarchicalAdversarialNetwork}
\end{figure*}

To overcome the above problems, we design a hierarchical adversarial network(HAN) for Chinese typeface transformation, consisting of a {\em transfer network} and a {\em hierarchical discriminator} (Fig. \ref{fig: HierarchicalAdversarialNetwork}), both of which are fully convolutional neural networks. First, different from existing {\em transfer network}, a staged-decoder is proposed which generates artificial images in multiple decoding layers, which is expected to help the decoder learn better representations in its hidden layers. Specially, the staged-decoder attempts to maximally preserve the global topological structure in different decoding layers simultaneously considers the local features decoded in hidden layers, thus enabling the transfer network to generate close to authentic characters instead of disordered strokes. Second, inspired by multi-classifier design in GoogLeNet~\cite{Szegedy2015Going}, which shows that the final feature layer may not provide rich and robust information for measuring the discrepancy between prediction and ground-truth, we propose a {\em hierarchical discriminator} for adversarial learning. Specifically, the discriminator introduce additional adversarial losses, each of which employs feature representations from different hidden layers. The multi-adversarial losses constitute a hierarchical form, enabling the discriminator to dynamically measure the discrepancy in distribution between the generated domain and target domain, so that the {\em transfer network} is trained to generate outputs with more similar statistical characteristics to the targets on different level of feature representation.
The main contribution of our work is summarized as follows.
\begin{itemize}
\setlength{\itemsep}{0pt}
\setlength{\parsep}{0pt}
\setlength{\parskip}{0pt}
    \item We introduce a staged-decoder in transfer network which generates multiple sets of characters based on different layers of decoded information, capturing both the global and local information for transfer. 
    \item We propose a {\em hierarchical discriminator} which involves a cascade of adversarial losses at different layers of the network, each providing complementary adversarial capability. We have experimentally shown that the hierarchical discriminator leads to faster model convergence and generates more realistic samples.
    \item The proposed hierarchical adversarial network(HAN) is shown to be successful for both typeface transfer and character restoration through extensive experimental studies. The impact of proposed hierarchical adversarial loss is further investigated from different perspective including gradient propagation and the ideology of adversarial training.
\end{itemize}


\section{Related Work}
\label{relatedwork}

Many natural image-to-image transformation tasks are domain transfer problem that maps images from source domain to target domain. This transformation can be formulated on pixel level(i.e. pixel-wise loss\cite{Zhang2016Colorful,Long2015Fully}) or more recently, feature level(i.e. perceptual loss~\cite{Johnson2016Perceptual}, Gram-Matrix~\cite{Gatys2016Image}, VGG loss~\cite{Ledig2016Photo}, style loss~\cite{Chen2017StyleBank}). Methods based on feature-level even can be extended to unsupervised situation in the assumption that both the input image and desired output image share identical or close high-level representations. However, this premise of assumption does not exist in handwriting transfer since sometimes the high-level representations between source characters and target ones are totally different.
Recently, generative adversarial networks~\cite{Goodfellow2014Generative}, especially its variants, CGAN~\cite{Mirza2014Conditional} and DCGAN~\cite{Radford2015Unsupervised}, have been successfully applied to a wide spectrum of image-to-image transformation tasks. Beyond the transfer network, CGAN-based methods introduce a discriminator, which involves an adversarial loss for constraining the distribution of generated domain to be close to that of target domain. The adversarial loss is employed by all the above GAN-based studies, such as image super-resolution~\cite{Ledig2016Photo}, de-noising~\cite{Zhang2017Image} and in-painting~\cite{Pathak2016Context}. Several studies leverage generator or discriminator to extract hidden-level representation and then perform feature matching in both domains~\cite{Dosovitskiy2016Generating,Wang2017Perceptual}.
\par In recent years, many image classification, detection or segmentation problems leverage the information in hidden layers of CNN for training. GoogLeNet~\cite{Szegedy2015Going} introduced auxiliary classifiers connected to intermediate layers based on the conclusion that the features produced by the layers in the middle of the network should also be very discriminative. Many other CNN models utilize features produced in different intermediate layers to construct extra loss function(~\cite{Wang2017Perceptual,Lyu2017Auto}). These auxiliary loss is thought to combat the gradient-vanish problem while providing regularization. We first applies this thought to discriminator in GAN, measuring the similarity of two distributions not only based on the high-level features but also relative low-lever features.


\section{Methods}
\label{methods}
In this section, we present the proposed Hierarchical Adversarial Network (HAN) for typeface transformation task. HAN consists of a {\em transfer network} and a {\em hierarchical discriminator}. The former is further consists of an Encoder and a Staged-Decoder. First, we introduce a {\em transfer network} $T$ which is responsible for mapping typeface-A characters to typeface-B characters. Then we introduce a hierarchical adversarial discriminator which helps the {\em transfer network} generate more realistic characters especially for the subtle structure in Chinese characters. Finally, we introduce the details of objective function.

\subsection{FCN-Based Transfer Network}
\label{sub:fcnbasedtransfernetwork}

\textbf{Encoder} The {\em transfer Network} has a similar architecture to that of ~\cite{Ronneberger2015U} with some modification. Because any information of relative-location is critical for Chinese character synthesis, we replace pooling operation with strided convolution in down-sampling since pooling helps reduce dimension and retains only robust activations in a receptive fields, however leading to the loss of spatial information in some degree. Additionally, it is a straightforward way to improve the performance of model by increasing the size of neural network, especially the depth. So more uniform-sized conv-layers were added in our encoder for extracting more local features (see Fig~\ref{fig: HierarchicalAdversarialNetwork}).

\textbf{Staged-Decoder}
Same as encoder, we insert additional uniform-sized convolution layers before each up-sampling conv-layer in decoder. A deeper decoder helps us model hierarchical representations of characters including the global topological structure and local topological of complicated Chinese characters. Considering the domain insight of our task in Section~\ref{introduction}.
We further propose a staged-decoder that leverages the hierarchical representation of decoder. Specifically, different intermediate features of decoder are utilized to generate characters ($T_1$, $T_2$) too. Together with the last generated characters ($T_3$), all of them will be sent to the discriminator(see Fig~\ref{fig: HierarchicalAdversarialNetwork}). We only measure the pixel-wise difference between the last generated characters ($T_3$) and corresponding ground-truth. The adversarial loss produced by $T_1$ and $T_2$ helps to refine the {\em transfer network}. Meanwhile, the loss produced by the intermediate layers of decoder may provide regularization for the parameters in transfer network, which will relieves the over-fitting problem in some degree. In addition, for typeface transformation, the input character and the desired output are expected to share underlying topological structure, but differ in appearance or style. Skip connection~\cite{Ronneberger2015U} is utilized to supplement partial invariant skeleton information of characters with encoded features concatenated on decoded features. Both encoder and staged decoder are fully convolutional networks~\cite{Long2015Fully}.

\subsection{Hierarchical Adversarial Discriminator} 
\label{sub:HierarchicalAdversarialDiscriminator}
As mentioned in Section~\ref{relatedwork}, adversarial loss introduced by discriminator is widely used in existing GAN-based image transformation task while all of them estimate the distribution consistency of two domain merely based on the final extracted features of discriminator. It is actually uncertain whether the learned features in last layers will provide rich and robust representations for discriminator. Additionally, We know the perceptual loss which penalizes the discrepancy between representations in different hidden space of images, is recently used in existing image-relative works. We combine the thought of perceptual loss and GANs, proposing a hierarchical adversarial discriminator which leverage the perceptual representations extracted from different intermediate layers of discriminator $D$ and then distinguishes real/fake distribution between generated domain $G_{domain}$ and target domain $T_{domain}$(See Fig~\ref{fig: HierarchicalAdversarialNetwork}). Each adversarial loss is defined as:
\begin{multline}
L_{d_{i}} = - \mathbb{E}_{f_{t}^{i}\sim p_{target}(f_{t})}[\log D_{i}(f_{t}^{i})] + \\ \mathbb{E}_{s\sim p_{source}(s)}[\log D_{i}(f_{s}^{i}(T(s))]
\label{eq: eq1}
\end{multline}
\begin{equation}
L_{g_{i}} = - \mathbb{E}_{s\sim p_{source}(s)}[\log D_{i}(f_{s}^{i}(T(s))]
\label{eq: eq2}
\end{equation}
where $f_t^{i}$ and $f_{s}^{i}(T(s))$ are $i^{th}$ perceptual representations learned in {\em Discriminator} from target domain and generated domain respectively. $D_i$ is branch discriminator cascaded after every intermediate layer and $i=1,2,..4$ which depends on the number of convolutional layers in our discriminator $D$.
This variation brings a complementary adversarial training for our model, which urges discriminator to find more detailed local discrepancy beyond the global distribution. Assuming $L_{d_{4}}$ and its corresponding $L_{g_{4}}$ reach nash equilibrium, which means the the perceptual representations $f_t^{4}$ and $f_{s}^{4}(T(s))$ are considered sharing the similar distribution, however other adversarial losses $($$L_{d_{i}}$, $L_{g_{i}}$$)$, $i\neq4$ may have not reach nash equilibrium since these losses produced by shallow losses pay more attention on regional information during training. The still high loss promotes the model to be continuously optimized until all perceptual representations pairs $($$f_t^{4}$, $f_{s}^{4}(T(s))$$)$, $i=1,2,..4$ are indistinguishable by discriminator. Experiments shows this strategy makes the discriminator to dynamically and automatically discover the un-optimized space from various perspectives.

Theoretically, our hierarchical adversarial discriminator actually plays an implicitly role of fitting distribution from two domains instead of fitting hidden features from paired images to be identical compared with existing methods. Thus our HAN model reduces the possibility of over-fitting and does not require pre-trained networks responsible for extracting features adopted by previous methods. Another merit our hierarchical adversarial strategy brought is that these auxiliary discriminators improve the flow of information and gradients throughout the network. The previous convolutional layers are optimized mainly by its neighbour adversarial loss beyond the other posterior adversarial losses so that the parameters existing in every discriminator layer is better optimized and the generator can thus be optimized better than before.

\subsection{Losses} 
\label{sub: Losses}

\textbf{Pixel-level Loss} The {\em transfer network} can be viewed as the generator in GANs. It aims to synthesize characters similar to the specified ground-truth ones. L1- or L2-norm are often used to measure the pixel distance between paired images. For our typeface transformation task, each pixel in character is normalized near 0 or 1 value. So cross entropy function is selected as per-pixel loss since this character generation problem can be viewed as a logistic regression. The pixel-wise loss is hence defined as follows:
\begin{multline}
    L_{pix-wise}(T) = \\ \mathbb{E}_{(s,t)}[-t\lambda_{w}\cdot(\log\sigma(T(s)))-(1-t)\cdot\log(1-\sigma(T(s)))],
\label{eq: eq3}
\end{multline}
where $T$ denotes the transformation of {\em transfer network}, $(s,t)$ is pair-wise samples where $s\sim p_{source\_domain}(s)$ and $t\sim p_{target\_domain}(t)$. $\sigma$ is $sigmoid$ activation.
Particularly, a weighted parameter $\lambda_{w}$ is introduced into pixel-wise loss for balancing the ratio of positive(value 0) to negative(value 1) pixels in every typeface style. We add this trade-off parameter based on the observation that some typefaces are thin (i.e. more negative pixels) while some may be relatively thick (i.e. more positive pixels). $\lambda_{w}$ is not a parameter determined by cross validation, it is explicitly defined by:
\begin{equation}
    \lambda_{w} = 1-\frac{\sum_{k=1}^{K}\sum_{n=1}^{N}\mathbbm{1}\left\{t_{k}^{n}\ge0.5\right\}}{\sum_{k=1}^{K}\sum_{n=1}^{N}\mathbbm{1}\left\{t_{k}^{n}<0.5\right\}},
\label{eq: eq4}
\end{equation}
where $N$ the is the resolution of one character image(here $N=64$), $K$ denotes the number of target characters in training set and $t_{k}^{n}$ denotes the $n^{th}$ pixel value of $k^{th}$ target character.

\textbf{Hierarchical Adversarial Loss}
For our proposed HAN, each adversarial loss is defined by Eq~\ref{eq: eq1} and Eq~\ref{eq: eq2}:
\begin{equation}
L_{adversarial}^{i}(D_{i},T) = L_{d_{i}} + L_{g_{i}}.
\label{eq: eq5}
\end{equation}
Noted that here we integrate original $t\sim p_{target}(t)$ and $s\sim p_{source}(s)$ into Eq.~\ref{eq: eq5} for a unified formulation, then the total adversarial losses is
\begin{equation}
L_{total\_adversarial}(D,T) = \sum_{i=1}^{k}\lambda_{i}\cdot L_{adversarial}^{i}(D_{i},T),
\label{eq: eq6}
\end{equation}
where $\lambda_{i}$ are weighted parameters to control the effect of every branch discriminator.
The total loss function is formulated as follows:
\begin{equation}
L_{total} = \lambda_{p} L_{pix-wise}(T) + \lambda_{a}L_{total\_adversarial}(D,T),
\label{eq: eq7}
\end{equation}
where  $\lambda_{p}$ and $\lambda_{a}$ are the trade-off parameters.

We optimize {\em transfer network} and {\em hierarchical adversarial discriminator} by turns.

\section{Experiments}
\label{experiments}

\subsection{Data Set}
\label{sub:datasetintroduction}
There is no public data set available for Chinese characters in different typefaces. We build a data set by downloading large amount of .ttf scripts denoting different typefaces from the the website \url{http://www.founder.com/}. After pre-processing, each typeface ends up with 6000+ grey-scale images in $64\times64$.png format. We choose a standard printed typeface named FangSong({\em FS}) as the source and the rest typefaces with handwriting styles are used as target ones. Most of our experiments use 50\% characters (\~3000 characters) as training set and the remaining as test set.

\subsection{Network Setup}
\label{sub:detailsetupofnetworkarchitecture}
The hyper-parameters relevant to our proposed network are annotated in Fig~\ref{fig: HierarchicalAdversarialNetwork}. The {\em encoder} includes 8 conv-layers while the {\em staged-Decoder} is more deeper including 4 transform-conv layers and 8 con-layers. Every $conv$ and $deconv$ are followed by Conv-BatchNorm(BN)~\cite{Ioffe2015Batch}-ELU~\cite{Clevert2015Fast}/ReLU structure. 4 skip connections are used on mirror layers both in {\em encoder} and {\em staged-decoder}.

For the trade-off parameters in Section~\ref{sub: Losses}, $\lambda_{w}$ is determined by Eq~\ref{eq: eq4}. The number of adversarial loss of HAN $l$ is 4 and weighted parameter $\{\lambda_{i}\}^{3}_{1}$ is decay from $1$ to $0.5$ with rate $0.9$, $\lambda_{4}=1.0$. $\lambda_{p}$ and $\lambda_{a}$ are both set to $1.0$ to weight the pixel loss and adversarial loss.

\subsection{Performance Comparison}
\label{sub:performancecomparison}
To validate the proposed HAN model, we compare the transfer performance of HAN with a Chinese calligraphy synthesis method (AEGG~\cite{Lyu2017Auto}) and two state-of-the-art image-to-image transformation methods(Pix2Pix~\cite{Isola2016Image}, Cycle-GAN~\cite{Zhu2017Unpaired}). Our proposed HAN can be trained in two modes. The first is {\em strong-paired} mode which minimizes pixel-wise discrepancy $L_{pix-wise}$ obtained by paired characters as well as hierarchical adversarial loss $L_{total_adversarial}$ obtained by generated and target domain. The second is {\em soft-paired} mode by removing $L_{pix-wise}$ and just minimizing $L_{total_adversarial}$, which looses the constrain of pairing source characters with corresponding target ones.

\begin{figure*}
\centering
\setlength{\abovecaptionskip}{-3pt}
\hspace{40mm}\textbf{\textit{Results on test-set}}
\hspace{50mm}\textbf{\textit{Results on train-set}}\\
\hspace{-10pt}
\subfigure{
\begin{minipage}{0.66\textwidth}{
\begin{minipage}{0.14\textwidth}\textit{Source}\end{minipage}
\begin{minipage}{0.45\textwidth}
\includegraphics[width=1.8\textwidth]{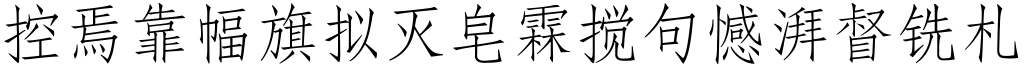}
\end{minipage}\\
\begin{minipage}{0.14\textwidth}\textit{AEGG}\end{minipage}
\begin{minipage}{0.45\textwidth}
\includegraphics[width=1.8\textwidth]{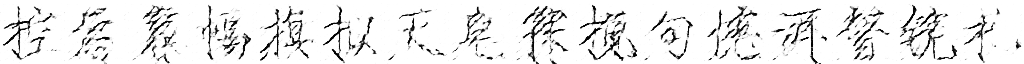}
\end{minipage}\\
\begin{minipage}{0.14\textwidth}\textit{Pix2Pix}\end{minipage}
\begin{minipage}{0.45\textwidth}
\includegraphics[width=1.8\textwidth]{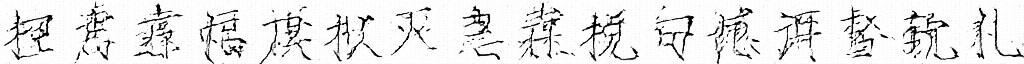}
\end{minipage}\\
\begin{minipage}{0.14\textwidth}\textbf{\textit{Ours}}\end{minipage}
\begin{minipage}{0.45\textwidth}
\includegraphics[width=1.8\textwidth]{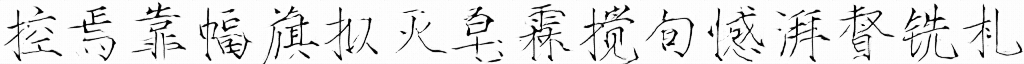}
\end{minipage}\\
\begin{minipage}{0.14\textwidth}\textit{Target}\end{minipage}
\begin{minipage}{0.45\textwidth}
\includegraphics[width=1.8\textwidth]{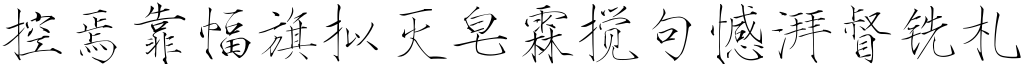}
\end{minipage}\\
}
\end{minipage}
}
\hspace{1pt}
\rulev
\hspace{3.5pt}
\subfigure{
\begin{minipage}{0.3\textwidth}{
\begin{minipage}{0.45\textwidth}
\includegraphics[width=2.0\textwidth]{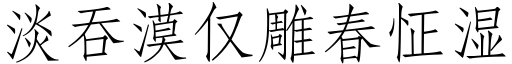}
\end{minipage}\\
\begin{minipage}{0.45\textwidth}
\includegraphics[width=2.0\textwidth]{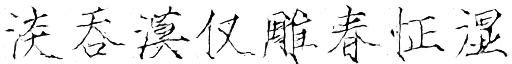}
\end{minipage}\\
\begin{minipage}{0.45\textwidth}
\includegraphics[width=2.0\textwidth]{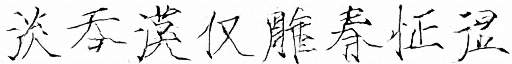}
\end{minipage}\\
\begin{minipage}{0.45\textwidth}
\includegraphics[width=2.0\textwidth]{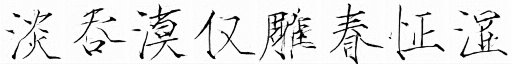}
\end{minipage}\\
\begin{minipage}{0.45\textwidth}
\includegraphics[width=2.0\textwidth]{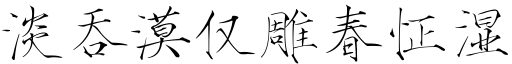}
\end{minipage}\\
}
\end{minipage}
}
\vspace{-10pt}
\ruleh
\hspace{-15pt}
\subfigure{
\begin{minipage}{0.66\textwidth}{
\begin{minipage}{0.14\textwidth}\textit{Source}\end{minipage}
\begin{minipage}{0.45\textwidth}
\includegraphics[width=1.8\textwidth]{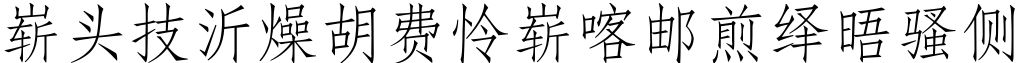}
\end{minipage}\\
\begin{minipage}{0.14\textwidth}\textit{AEGG}\end{minipage}
\begin{minipage}{0.45\textwidth}
\includegraphics[width=1.8\textwidth]{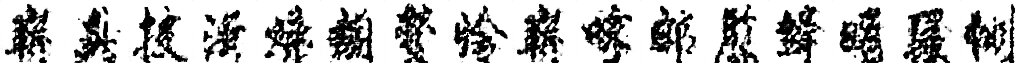}
\end{minipage}\\
\begin{minipage}{0.14\textwidth}\textit{Pix2Pix}\end{minipage}
\begin{minipage}{0.45\textwidth}
\includegraphics[width=1.8\textwidth]{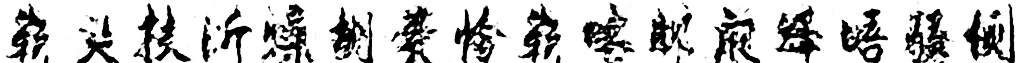}
\end{minipage}\\
\begin{minipage}{0.14\textwidth}\textbf{\textit{Ours}}\end{minipage}
\begin{minipage}{0.45\textwidth}
\includegraphics[width=1.8\textwidth]{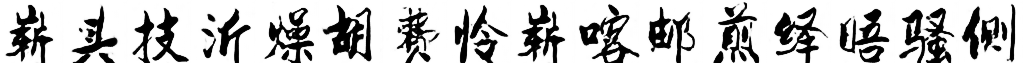}
\end{minipage}\\
\begin{minipage}{0.14\textwidth}\textit{Target}\end{minipage}
\begin{minipage}{0.45\textwidth}
\includegraphics[width=1.8\textwidth]{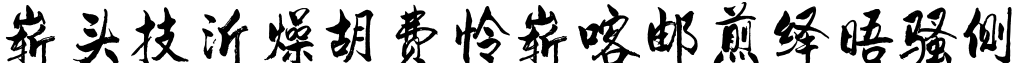}
\end{minipage}\\
}
\end{minipage}
}
\hspace{1pt}
\rulev
\hspace{1pt}
\subfigure{
\begin{minipage}{0.3\textwidth}{
\begin{minipage}{0.45\textwidth}
\includegraphics[width=2.0\textwidth]{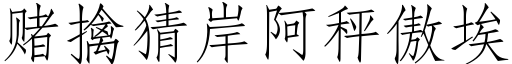}
\end{minipage}\\
\begin{minipage}{0.45\textwidth}
\includegraphics[width=2.0\textwidth]{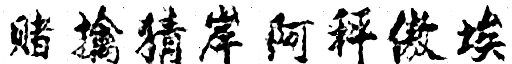}
\end{minipage}\\
\begin{minipage}{0.45\textwidth}
\includegraphics[width=2.0\textwidth]{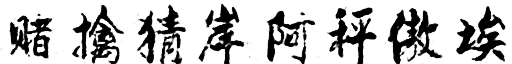}
\end{minipage}\\
\begin{minipage}{0.45\textwidth}
\includegraphics[width=2.0\textwidth]{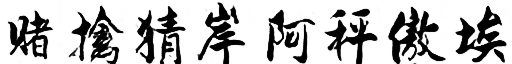}
\end{minipage}\\
\begin{minipage}{0.45\textwidth}
\includegraphics[width=2.0\textwidth]{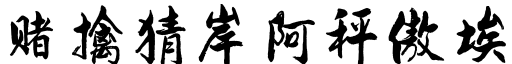}
\end{minipage}\\
}
\end{minipage}
}
\vspace{-10pt}
\ruleh
\hspace{-15pt}
\subfigure{
\begin{minipage}{0.66\textwidth}{
\begin{minipage}{0.14\textwidth}\textit{Source}\end{minipage}
\begin{minipage}{0.45\textwidth}
\includegraphics[width=1.8\textwidth]{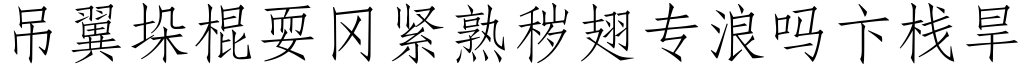}
\end{minipage}\\
\begin{minipage}{0.14\textwidth}\textit{AEGG}\end{minipage}
\begin{minipage}{0.45\textwidth}
\includegraphics[width=1.8\textwidth]{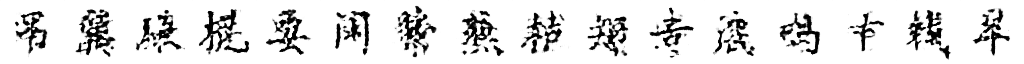}
\end{minipage}\\
\begin{minipage}{0.14\textwidth}\textit{Pix2Pix}\end{minipage}
\begin{minipage}{0.45\textwidth}
\includegraphics[width=1.8\textwidth]{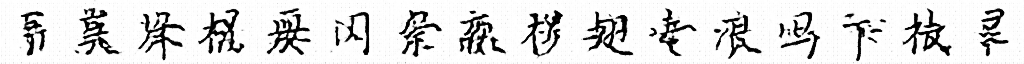}
\end{minipage}\\
\begin{minipage}{0.14\textwidth}\textbf{\textit{Ours}}\end{minipage}
\begin{minipage}{0.45\textwidth}
\includegraphics[width=1.8\textwidth]{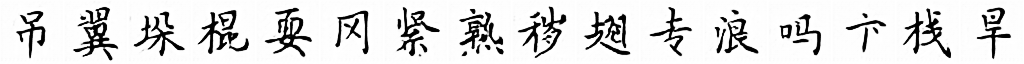}
\end{minipage}\\
\begin{minipage}{0.14\textwidth}\textit{Target}\end{minipage}
\begin{minipage}{0.45\textwidth}
\includegraphics[width=1.8\textwidth]{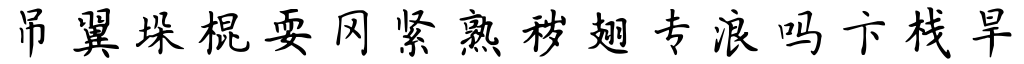}
\end{minipage}\\
}
\end{minipage}
}
\hspace{1pt}
\rulev
\hspace{1pt}
\subfigure{
\begin{minipage}{0.3\textwidth}{
\begin{minipage}{0.45\textwidth}
\includegraphics[width=2.0\textwidth]{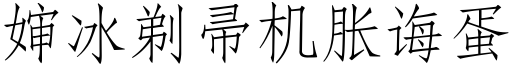}
\end{minipage}\\
\begin{minipage}{0.45\textwidth}
\includegraphics[width=2.0\textwidth]{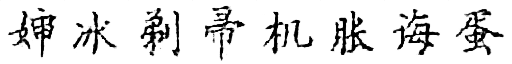}
\end{minipage}\\
\begin{minipage}{0.45\textwidth}
\includegraphics[width=2.0\textwidth]{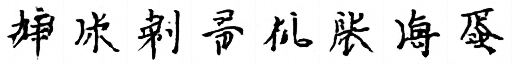}
\end{minipage}\\
\begin{minipage}{0.45\textwidth}
\includegraphics[width=2.0\textwidth]{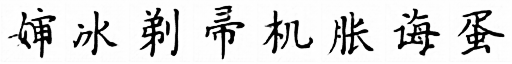}
\end{minipage}\\
\begin{minipage}{0.45\textwidth}
\includegraphics[width=2.0\textwidth]{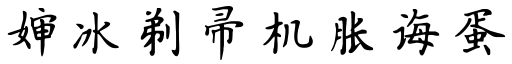}
\end{minipage}\\
}
\end{minipage}
}
\vspace{-10pt}
\ruleh
\hspace{-15pt}
\subfigure{
\begin{minipage}{0.66\textwidth}{
\begin{minipage}{0.14\textwidth}\textit{Source}\end{minipage}
\begin{minipage}{0.45\textwidth}
\includegraphics[width=1.8\textwidth]{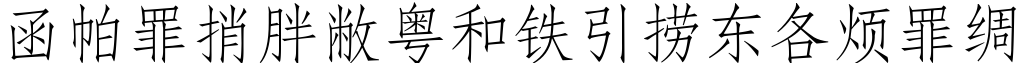}
\end{minipage}\\
\begin{minipage}{0.14\textwidth}\textit{AEGG}\end{minipage}
\begin{minipage}{0.45\textwidth}
\includegraphics[width=1.8\textwidth]{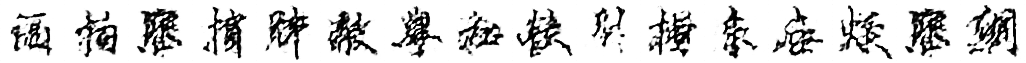}
\end{minipage}\\
\begin{minipage}{0.14\textwidth}\textit{Pix2Pix}\end{minipage}
\begin{minipage}{0.45\textwidth}
\includegraphics[width=1.8\textwidth]{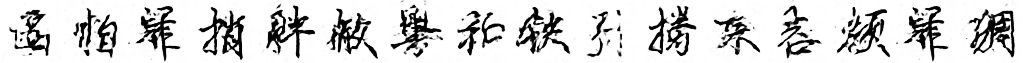}
\end{minipage}\\
\begin{minipage}{0.14\textwidth}\textbf{\textit{Ours}}\end{minipage}
\begin{minipage}{0.45\textwidth}
\includegraphics[width=1.8\textwidth]{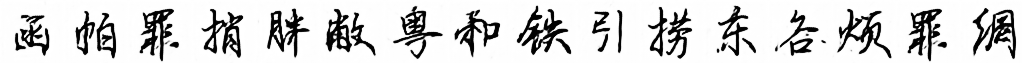}
\end{minipage}\\
\begin{minipage}{0.14\textwidth}\textit{Target}\end{minipage}
\begin{minipage}{0.45\textwidth}
\includegraphics[width=1.8\textwidth]{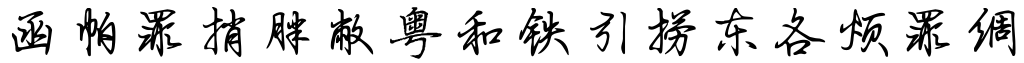}
\end{minipage}\\
}
\end{minipage}
}
\hspace{1pt}
\rulev
\hspace{1pt}
\subfigure{
\begin{minipage}{0.3\textwidth}{
\begin{minipage}{0.45\textwidth}
\includegraphics[width=2.0\textwidth]{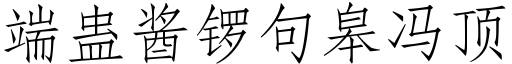}
\end{minipage}\\
\begin{minipage}{0.45\textwidth}
\includegraphics[width=2.0\textwidth]{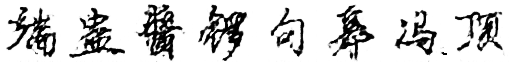}
\end{minipage}\\
\begin{minipage}{0.45\textwidth}
\includegraphics[width=2.0\textwidth]{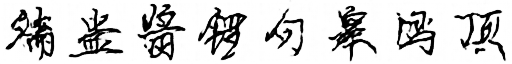}
\end{minipage}\\
\begin{minipage}{0.45\textwidth}
\includegraphics[width=2.0\textwidth]{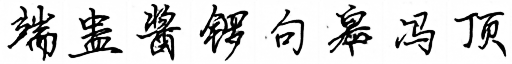}
\end{minipage}\\
\begin{minipage}{0.45\textwidth}
\includegraphics[width=2.0\textwidth]{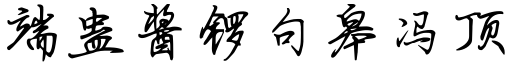}
\end{minipage}\\
}
\end{minipage}
}
\vspace{-10pt}
\ruleh
\hspace{-15pt}
\subfigure{
\begin{minipage}{0.66\textwidth}{
\begin{minipage}{0.14\textwidth}\textit{Source}\end{minipage}
\begin{minipage}{0.45\textwidth}
\includegraphics[width=1.8\textwidth]{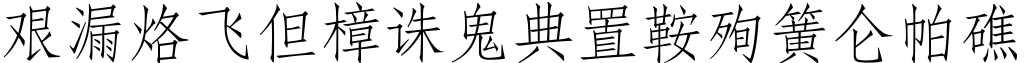}
\end{minipage}\\
\begin{minipage}{0.14\textwidth}\textit{AEGG}\end{minipage}
\begin{minipage}{0.45\textwidth}
\includegraphics[width=1.8\textwidth]{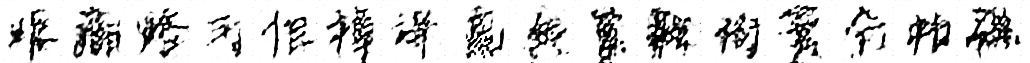}
\end{minipage}\\
\begin{minipage}{0.14\textwidth}\textit{Pix2Pix}\end{minipage}
\begin{minipage}{0.45\textwidth}
\includegraphics[width=1.8\textwidth]{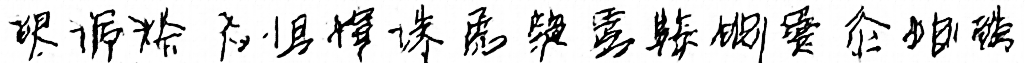}
\end{minipage}\\
\begin{minipage}{0.14\textwidth}\textbf{\textit{Ours}}\end{minipage}
\begin{minipage}{0.45\textwidth}
\includegraphics[width=1.8\textwidth]{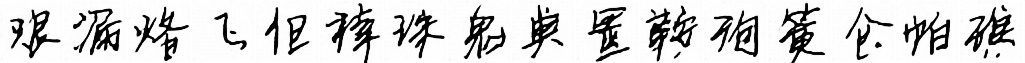}
\end{minipage}\\
\begin{minipage}{0.14\textwidth}\textit{Target}\end{minipage}
\begin{minipage}{0.45\textwidth}
\includegraphics[width=1.8\textwidth]{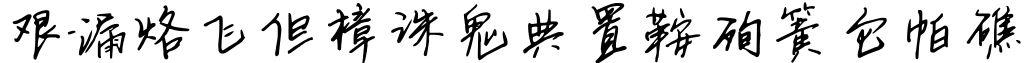}
\end{minipage}\\
}
\end{minipage}
}
\hspace{1pt}
\rulev
\hspace{3.5pt}
\subfigure{
\begin{minipage}{0.3\textwidth}{
\begin{minipage}{0.45\textwidth}
\includegraphics[width=2.0\textwidth]{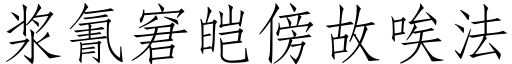}
\end{minipage}\\
\begin{minipage}{0.45\textwidth}
\includegraphics[width=2.0\textwidth]{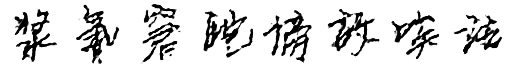}
\end{minipage}\\
\begin{minipage}{0.45\textwidth}
\includegraphics[width=2.0\textwidth]{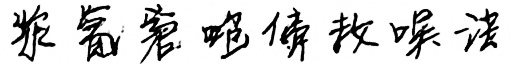}
\end{minipage}\\
\begin{minipage}{0.45\textwidth}
\includegraphics[width=2.0\textwidth]{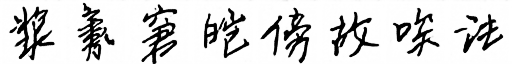}
\end{minipage}\\
\begin{minipage}{0.45\textwidth}
\includegraphics[width=2.0\textwidth]{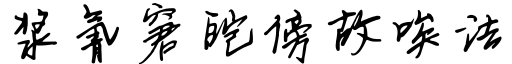}
\end{minipage}\\
}
\end{minipage}
}
\ruleh
\hspace{60mm}\textbf{Transfer {\em FS}-typeface to 5 personal {\em handwriting}-styles typeface}
\ruleh
\caption{Performance of transferring {\em FS} typeface to other 5 personal {\em handwriting}-style typefaces.}
\label{fig: ExperimentalPerformance}
\end{figure*}

\noindent \textbf{Strong-Paired Learning.}
Baseline AEGG and Pix2Pix both need pair the generated images with corresponding ground-truths for training so we compare our HAN with them in {\em strong-paired} mode. The {\em transfer network} of Pix2Pix shares the identical framework with that in our HAN(see Fig~\ref{fig: HierarchicalAdversarialNetwork}) and the model used in AEGG follows the instructions of their paper with some tiny adjustment for dimension adaptation. 50\%(\~3000) characters randomly selected from {\em FS} typeface as well as 50\% corresponding target style characters selected from other handwriting-style typeface are used as training set. The remaining 50\% of {\em FS} typefaces is used for testing. We perform 5 experiments transferring {\em FS} typeface to other Chinese handwriting-style(see Fig~\ref{fig: ExperimentalPerformance}). All methods can capture general style of handwriting however AEGG and Pix2Pix failed to synthesize recognizable characters because most strokes in generated character are disordered even chaotic. Our HAN significantly outperforms AEGG and Pix2Pix, especially imitating cursive handwriting characters. Experimental result shows HAN is superior in generating detailed component of characters. We also observed that both baselines perform well on training set but far worse on test set, which suggests the proposed hierarchical adversarial loss makes our model less prone to over-fitting in some degree.

\begin{figure*}
\centering
\subfigure{
	\begin{minipage}{0.15\textwidth}\textit{Source Characters}\end{minipage}
}
\hspace{1.0pt}
\rulev
\hspace{1pt}
\subfigure{
	\begin{minipage}{0.2\textwidth}
	\includegraphics[width=0.8\textwidth]{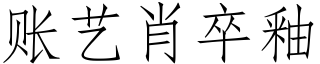}
	\end{minipage}
}
\hspace{-20pt}
\rulev
\hspace{1pt}
\subfigure{
	\begin{minipage}{0.2\textwidth}
	\includegraphics[width=0.8\textwidth]{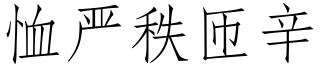}
	\end{minipage}
}
\hspace{-20pt}
\rulev
\hspace{1pt}
\subfigure{
	\begin{minipage}{0.2\textwidth}
	\includegraphics[width=0.8\textwidth]{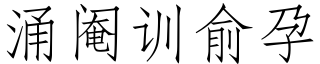}
	\end{minipage}
}
\hspace{-20pt}
\rulev
\hspace{1pt}
\subfigure{
	\begin{minipage}{0.2\textwidth}
	\includegraphics[width=0.8\textwidth]{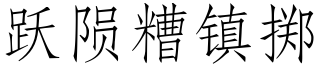}
	\end{minipage}
}
\ruleh
\hspace{-5pt}
\subfigure{
	\begin{minipage}{0.20\textwidth}{
	\vspace{2mm}
	\begin{minipage}{0.8\textwidth}\textit{HAN(strong-pair)}
	\vspace{4mm}
	\end{minipage}\\
	\begin{minipage}{0.8\textwidth}\textit{CycleG(soft-pair)}
	\vspace{4mm}
	\end{minipage}\\
	\begin{minipage}{0.8\textwidth}\textbf{\textit{HAN(soft-pair)}}
	\end{minipage}\\
	}
	\end{minipage}
}
\hspace{-24pt}
\rulev
\hspace{1pt}
\subfigure{
	\begin{minipage}{0.2\textwidth}{
	\begin{minipage}{1.0\textwidth}	
	\includegraphics[width=0.85\textwidth]{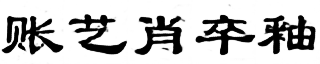}
	\vspace{1mm}
	\end{minipage}\\
	\begin{minipage}{1.0\textwidth}	
	\includegraphics[width=0.85\textwidth]{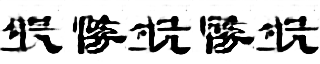}
	\vspace{1mm}
	\end{minipage}\\	
	\begin{minipage}{1.0\textwidth}	
	\includegraphics[width=0.85\textwidth]{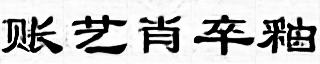}
	\end{minipage}\\
	}
	\end{minipage}
}
\hspace{-20pt}
\rulev
\hspace{1pt}
\subfigure{
	\begin{minipage}{0.2\textwidth}{
	\begin{minipage}{1.0\textwidth}	
	\includegraphics[width=0.85\textwidth]{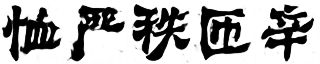}
	\vspace{1mm}
	\end{minipage}\\
	\begin{minipage}{1.0\textwidth}	
	\includegraphics[width=0.85\textwidth]{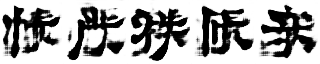}
	\vspace{1mm}
	\end{minipage}\\	
	\begin{minipage}{1.0\textwidth}	
	\includegraphics[width=0.85\textwidth]{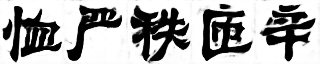}
	\end{minipage}\\
	}
	\end{minipage}
}
\hspace{-20pt}
\rulev
\hspace{1pt}
\subfigure{
	\begin{minipage}{0.2\textwidth}{
	\begin{minipage}{1.0\textwidth}	
	\includegraphics[width=0.85\textwidth]{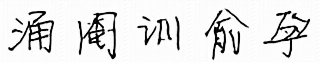}
	\vspace{1mm}
	\end{minipage}\\
	\begin{minipage}{1.0\textwidth}	
	\includegraphics[width=0.85\textwidth]{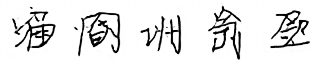}
	\vspace{1mm}
	\end{minipage}\\	
	\begin{minipage}{1.0\textwidth}	
	\includegraphics[width=0.85\textwidth]{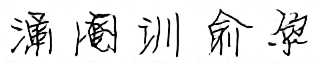}
	\end{minipage}\\
	}
	\end{minipage}
}
\hspace{-20pt}
\rulev
\hspace{1pt}
\subfigure{
	\begin{minipage}{0.2\textwidth}{
	\begin{minipage}{1.0\textwidth}	
	\includegraphics[width=0.85\textwidth]{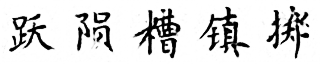}
	\vspace{1mm}
	\end{minipage}\\
	\begin{minipage}{1.0\textwidth}	
	\includegraphics[width=0.85\textwidth]{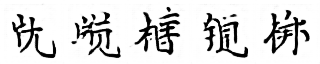}
	\vspace{1mm}
	\end{minipage}\\	
	\begin{minipage}{1.0\textwidth}	
	\includegraphics[width=0.85\textwidth]{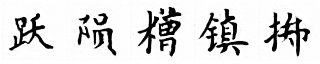}
	\end{minipage}\\
	}
	\end{minipage}
}
\ruleh
\hspace{-4pt}
\subfigure{
	\begin{minipage}{0.15\textwidth}\textit{Target Characters}\end{minipage}
}
\hspace{0pt}
\rulev
\hspace{1pt}
\subfigure{
	\begin{minipage}{0.2\textwidth}
	\includegraphics[width=0.8\textwidth]{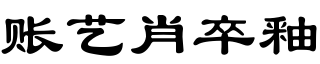}
	\end{minipage}
}
\hspace{-20pt}
\rulev
\hspace{1pt}
\subfigure{
	\begin{minipage}{0.2\textwidth}
	\includegraphics[width=0.8\textwidth]{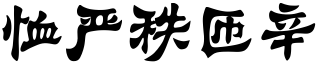}
	\end{minipage}
}
\hspace{-20pt}
\rulev
\hspace{1pt}
\subfigure{
	\begin{minipage}{0.2\textwidth}
	\includegraphics[width=0.8\textwidth]{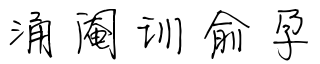}
	\end{minipage}
}
\hspace{-20pt}
\rulev
\hspace{1pt}
\subfigure{
	\begin{minipage}{0.2\textwidth}
	\includegraphics[width=0.8\textwidth]{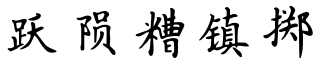}
	\end{minipage}
}
\caption{ We compare our HAN with Cycle-GAN by loosing the pairing constraint and HAN performs better than Cycle-GAN.}
\label{fig: SoftPairedPerformance}
\end{figure*}

\noindent \textbf{Soft-Paired Learning.}
Another model Cycle-GAN actually is an unpaired method which does not require ground-truth for training. Nevertheless we experiment unpaired form with Cycle-GAN and proposed HAN, both of their results are very bad. So we compare our HAN with Cycle-GAN in {\em soft-paired} mode, saving the trouble of tedious pairing but leaving the ground-truths in training set. As illustrated in Fig~\ref{fig: SoftPairedPerformance}, under the condition of {\em soft-paired}, our HAN performs well than Cycle-GAN. Though Cycle-GAN correctly captures the style of target characters, it cannot reconstruct correct location of every stroke and Cycle-GAN leads to model collapse. Of course, results of HAN trained in {\em soft-paired} mode is not as good as that {\em strong-paired} mode since the strong supervision information is reduced by removing $L_{pix-wise}$.

\noindent \textbf{Quantitative Evaluation.}
Beyond directly illustrating qualitative results of comparison experiments, two quantitative measurements: Root Mean Square Error(RMSE) and Average Pixel Disagreement Ration~\cite{Liu2016Coupled}(APDR) are utilized as evaluation criterion. As shown in Table~\ref{tab: QuantitativeMeasurements}, our HAN leads to the lowest RMSE and APDR value both under the mode of {\em strong-paired} and {\em soft-paired} mode compared with existing methods.
\begin{table*}[htb]
\centering
\begin{threeparttable}
	\begin{tabular}{lcccccccc}
	\toprule[1pt]
	\multirow{2}{*}{Model}&
	\multicolumn{2}{c}{$FS$$\to$$handwriting1$}&\multicolumn{2}{c}{$FS$$\to$$handwriting2$}&\multicolumn{2}{c}{$FS$$\to$$handwriting3$}&\multicolumn{2}{c}{$FS$$\to$$handwriting4$}\\
	\cmidrule(lr){2-3} \cmidrule(lr){4-5} \cmidrule(lr){6-7} \cmidrule(lr){8-9}
	&RMSE&APDR&RMSE&APDR&RMSE&APDR&RMSE&APDR\\
	\midrule
	AEGG~\cite{Lyu2017Auto}  &22.671&0.143 &28.010&0.211 &24.083&0.171 &22.110&0.131\\
	\rowcolor[gray]{0.9} Pix2Pix~\cite{Isola2016Image} &29.731&0.231 &27.117&0.225 &26.580&0.187 &24.135&0.180\\
	Cycle-GAN~\cite{Zhu2017Unpaired}  &29.602&0.253  &29.145&0.234  &28.845&0.241 &25.632&0.191 \\
	\rowcolor[gray]{0.9}\textbf{HAN(Soft-pair)}   &20.984&0.125 &25.442&0.207 &24.741&0.181 &20.714&0.134\\
	\textbf{HAN(Strong-pair)}  &\textbf{19.498}&\textbf{0.118} &\textbf{23.303}&\textbf{0.181} &\textbf{22.266}&\textbf{0.162} &\textbf{19.528}&\textbf{0.110}\\
	\bottomrule[1pt]
	\end{tabular}
\caption{Quantitative Measurements}
\label{tab: QuantitativeMeasurements}
\end{threeparttable}
\end{table*}

\subsection{Analysis of Hierarchical Adversarial Loss}
\label{sub:hanVersusSAN}

\begin{figure}[htb]
\centering
\includegraphics[width=\linewidth,height=5.0cm]{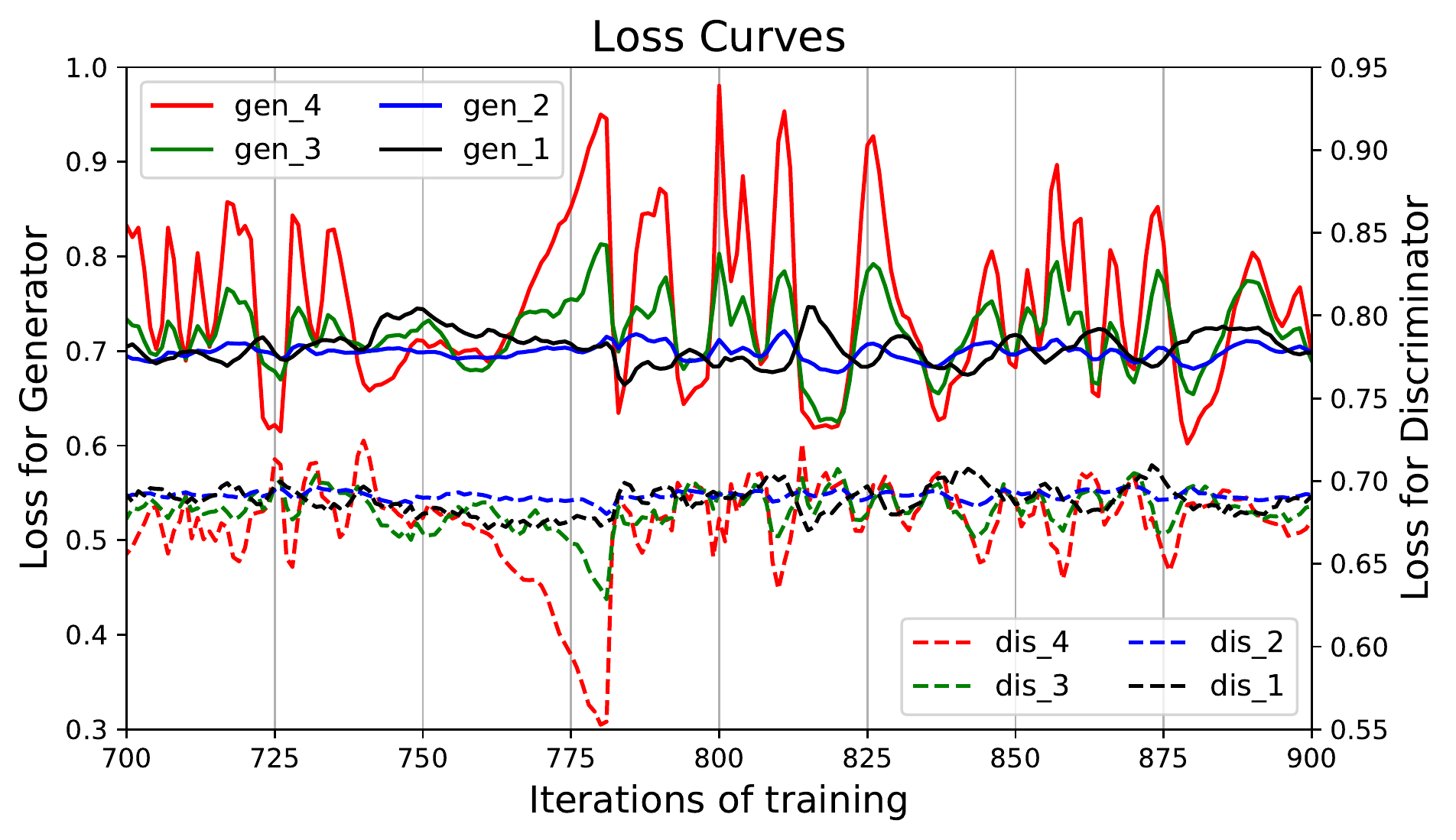}
\caption{Each generator loss and discriminator loss during ste 700 to 900.}
\label{fig: hierarchicaladversarialloss}
\end{figure}

We analyze each adversarial loss, $\{L_{d_{i}}\}^{4}_{i=1}$ and $\{L_{g_{i}}\}^{4}_{i=1}$, defined in Section~\ref{sub:HierarchicalAdversarialDiscriminator}. As shown in Fig~\ref{fig: hierarchicaladversarialloss}, the generator loss $gen\_4$ produced by the last conv-layer in hierarchical discriminator fluctuates greatly and then $gen\_3$ produced by the penultimate layer, $\{gen\_2, gen\_1\}$ produced by shallower conv-layers are relatively gentle because $\lambda_4$ is set larger than $\{\lambda_{i}\}^{3}_{i=1}$ so that network mainly optimizes $gen\_4$. However for discriminator loss, $\{dis\_4, dis\_3, dis\_1\}$ derived from $D_4$, $D_3$,$D_1$ are mostly numerical approach. We further observed that the trend of increase or reduction among various discriminator losses are not always consistent. We experimentally conclude that adversarial losses produced by intermediate layers can assist training: when $D_4$ is severely cheated by real/fake characters, $D_3$ or $D_2$ or $D_1$ can still give a high confidence of differentiating, which means True/False discrimination based on different representations can be compensated each other(see Fig~\ref{fig: hierarchicaladversarialloss} for more details) during training.

\begin{figure*}[htb]
\centering
\includegraphics[width=1.0\textwidth]{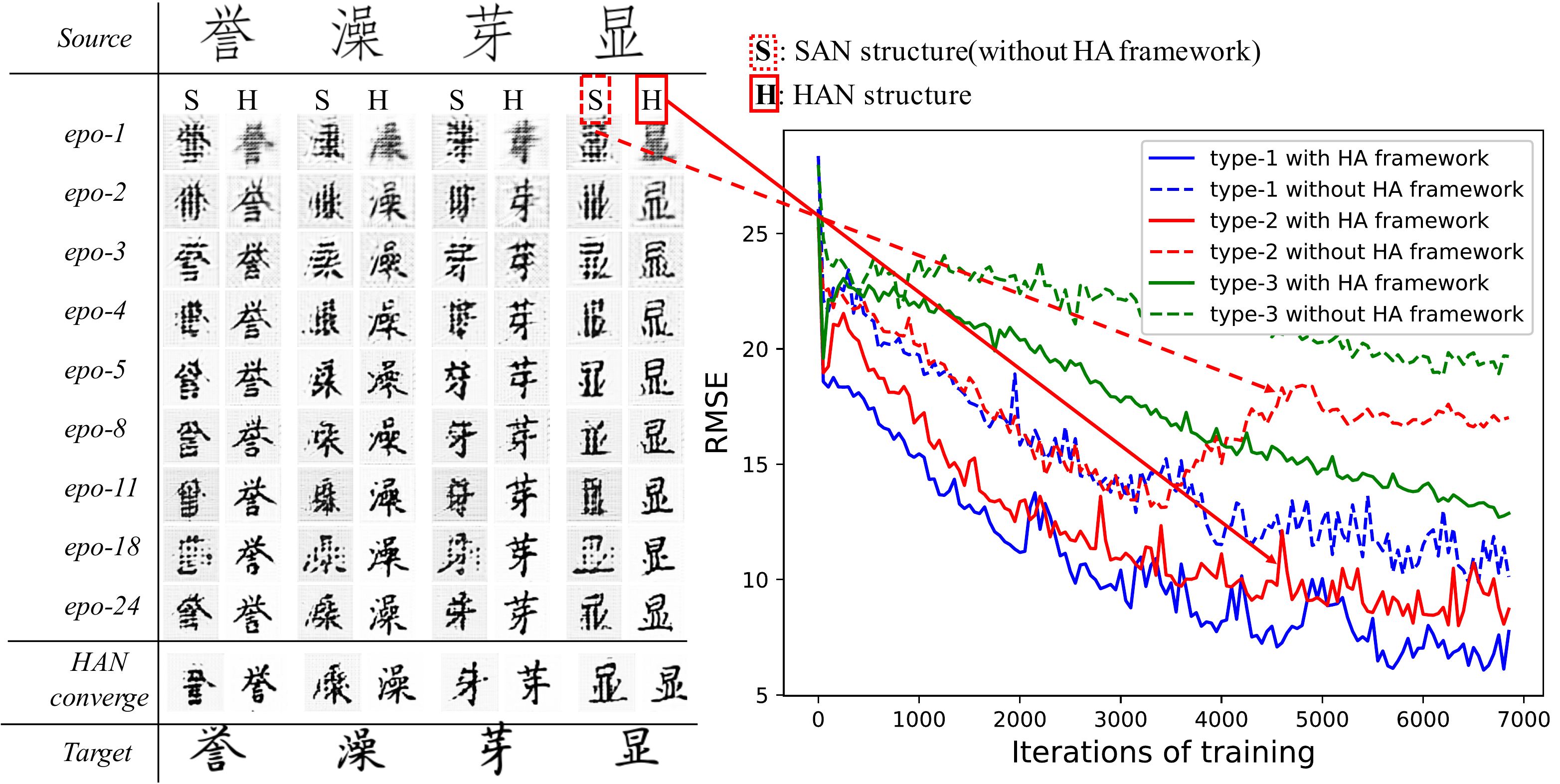}
\caption{Contrast experiments for HAN and SAN. Characters generated by HAN are far more better than that by SAN in same training epoch. {\em HAN converge} row shows characters generated  when our HAN model converges. The RMSE evaluation loss along with the training iterations under HAN and SAN shows HAN leads to more lower value than SAN.}
\label{fig: comparisonbetweenHAorNonHA}
\end{figure*}

We further explore the influence brought by our hierarchical adversarial loss. By removing the effect of hierarchical architecture from our HAN model, we run another contrast experiment, Single Adversarial Network (SAN). The detail of network follows Fig~\ref{fig: HierarchicalAdversarialNetwork} and we set trade-off parameters $\lambda_1=\lambda_2=\lambda_3=0.5$ and $\lambda_4=1$ in loss function of HAN, while we set $\lambda_1=\lambda_2=\lambda_3=0$ and $\lambda_4=1$ for SAN in order to remove the influence of extra 3 adversarial losses. Considering the value of hierarchical adversarial loss(we accumulate four adversarial losses) is bigger than that of single adversarial loss,  the gradients in back propagation of HAN is hence theoretically bigger than that of SAN. For demonstrating that our HAN works not for this reason, we multiply a constant $c=\frac{\lambda_1+\lambda_2+\lambda_3+\lambda_4}{\lambda_4}$ before the adversarial loss in SAN so that these two adversarial loss respectively existing in HAN and SAN are close proximity. Characters generated during different training period are illustrated in Fig~\ref{fig: comparisonbetweenHAorNonHA} from which we can see qualitative effect of proposed hierarchical adversarial discriminator. Our proposed HAN generates more clear characters compared with SAN at the same phase of training period, which suggests HAN converge greatly faster than SAN. We also run 3 parallel typeface-transfer experiments then calculate RMSE along with the iterations of training on train set. Left loss-curves in Fig~\ref{fig: comparisonbetweenHAorNonHA} demonstrates that hierarchical adversarial architecture assists to accelerate convergence and leads to lower RMSE value.

\subsection{Character Restoration with HAN}
\begin{figure}
\centering
    \includegraphics[width=0.48\textwidth]{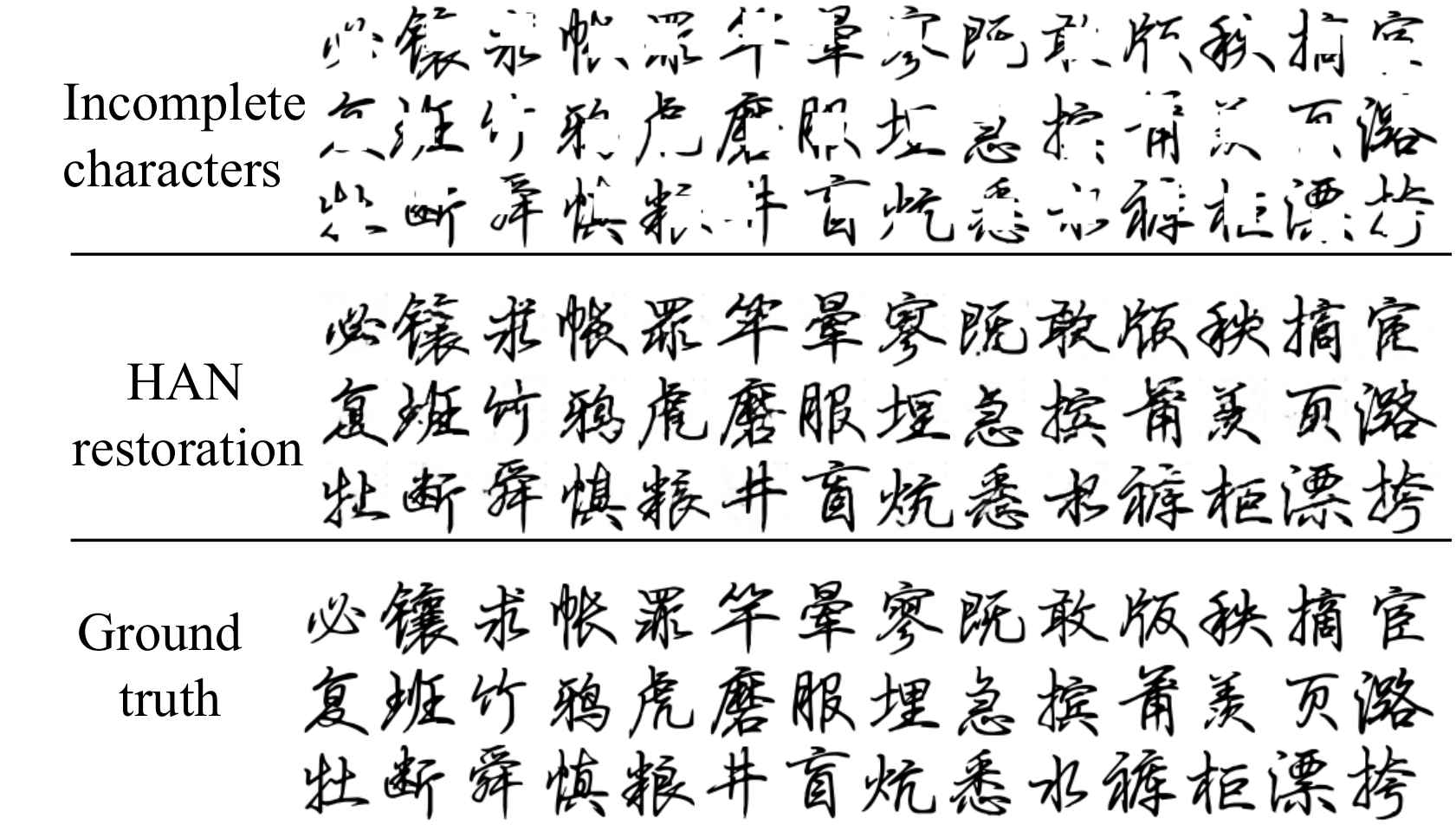}
\caption{Performance of repairing personal handwriting characters with HAN on test set.}
\label{fig: characterrestoration}
\end{figure}

Beyond transferring standard printed typeface to any handwriting-style typeface, we also applied our HAN model to character restoration. We randomly mask 30\% region on every handwriting characters in one typeface's training set. Under {\em strong-paired} mode, our HAN learned to correctly reconstruct the original characters. As illustrated in Fig~\ref{fig: characterrestoration}, our HAN is able to correctly reconstruct the missing part of one character on test set.

\subsection{Impact of Training Set Size}
\label{sub:differenttrainingsamples}
\begin{figure}[htb]
\centering
\includegraphics[width=\linewidth, height=0.30\textwidth]{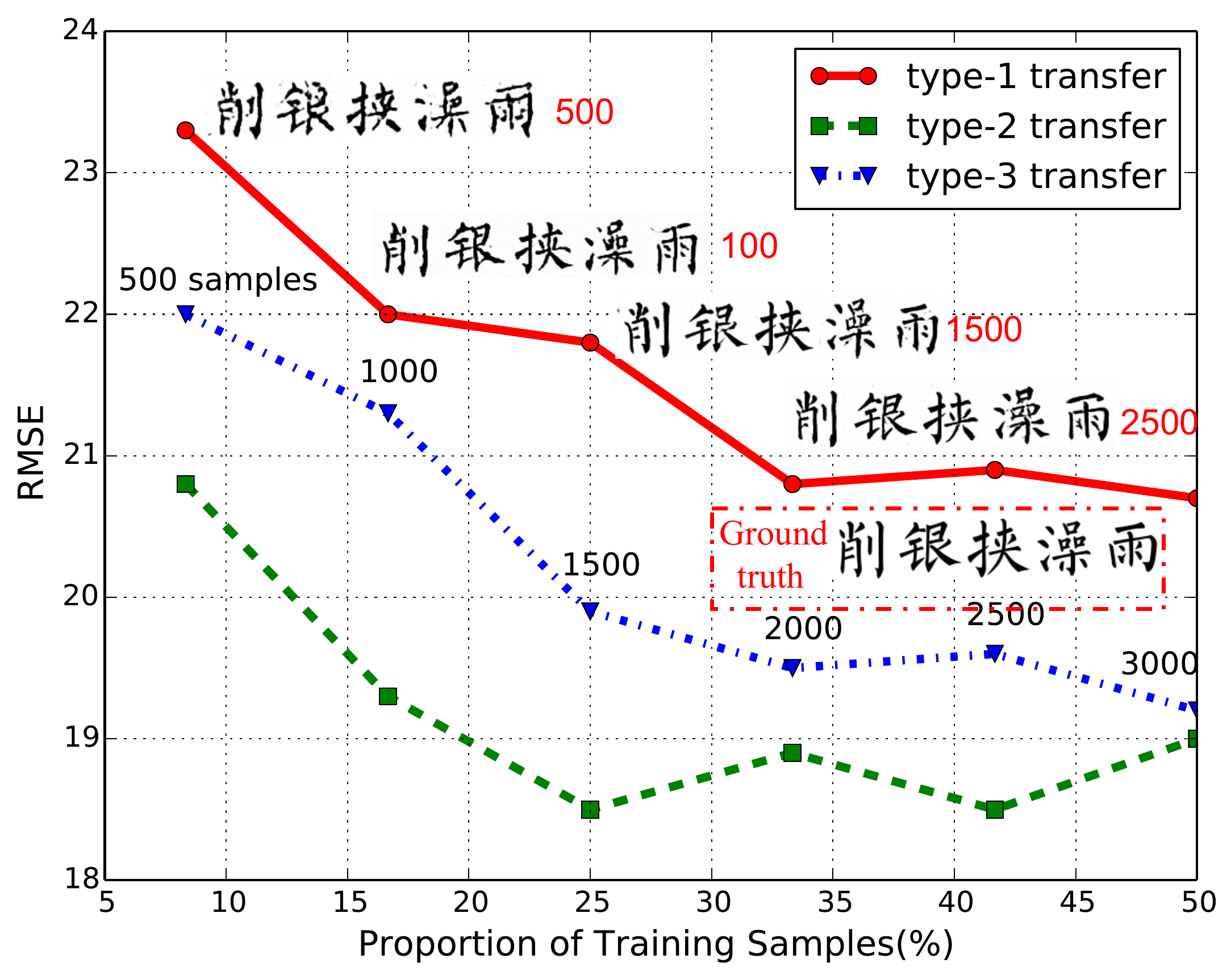}
\caption{The RMSE evaluation under different proportion of training set. The red and black number denote how many train samples we used. We present 3 transferring handwriting characters from {\em FS}-typeface to handwriting type-1, type-2 and type-3.}
\label{fig: rmse}
\end{figure}

Last, we experiment at least how many handwriting characters should be given in training to ensure a satisfied transfer performance. So we experiment three typeface-transfer tasks(type-1, type-2 and type-3) with different proportion of training samples and then evaluate on each test set. As the synthesized characters shown in Fig~\ref{fig: rmse}, the performance improves along with increase of training samples. We also use RMSE to quantify the performance under different training samples. All 3 curves suggests when the proportion of training size is not less than 35\%(2000 samples), the performance will not be greatly improved.

\section{Conclusion and Future Work}
\label{conclusionandfuturework}

In this paper, we propose a hierarchical adversarial network (HAN) for typeface transformation. The HAN is consisted of a {\em transfer network} and a {\em hierarchical adversarial discriminator}. The {\em transfer network} consists of an encoder and a staged-decoder which can generate characters based on different decoded information. The proposed {\em hierarchical discriminator} can dynamically estimate the consistency of two domains from different-level perceptual representations, which helps our HAN converge faster and better. Experimental results show our HAN can synthesize most handwriting-style typeface compared with existing natural image-to-image transformation methods. Additionally, our HAN can be applied to handwriting character restoration.


{\small
\bibliographystyle{ieee}
\bibliography{egbib}
}

\end{document}